\documentclass[journal]{IEEEtran}

\usepackage{multirow}

\ifCLASSINFOpdf
  \usepackage[pdftex]{graphicx}
\else
\fi

\usepackage{amsmath}
\usepackage{array}
\usepackage{algorithm}
\usepackage{algorithmic}
\usepackage{cite}
\usepackage{bm}
\usepackage{caption}
\usepackage{hyperref}
\usepackage{amssymb}
\usepackage{color,xcolor}
\usepackage{subfigure}

\usepackage{ragged2e}

\begin{document}
%
\title{Uncertainty-Aware Collaborative System of Large and Small Models for Multimodal Sentiment Analysis}
%
%
%

\author{Shiqin Han,
        Manning Gao,
        Menghua Jiang,
        Yuncheng Jiang,
        Haifeng Hu,
        Sijie Mai\IEEEauthorrefmark{1}
\IEEEcompsocitemizethanks{
\IEEEcompsocthanksitem * Corresponding Author.
\IEEEcompsocthanksitem Shiqin Han, Menghua Jiang, Yuncheng Jiang, Sijie Mai are with School of Computer Science, South China Normal University, China. 
E-mail: 20222121019@m.scnu.edu.cn, jiangmenghua@m.scnu.edu.cn, ycjiang@scnu.edu.cn, sijiemai@m.scnu.edu.cn
\IEEEcompsocthanksitem Manning Gao is with School of Artificial Intelligence, South China Normal University, China.
E-mail: 20232005149@m.scnu.edu.cn
\IEEEcompsocthanksitem Haifeng Hu is with School of Electronics and Information Technology, Sun Yat-sen University, China. 
E-mail: huhaif@mail.sysu.edu.cn
}
\thanks{}
}

\markboth{IEEE TRANSACTIONS ON AFFECTING COMPUTING}%
{Shell \MakeLowercase{\textit{et al.}}: Bare Demo of IEEEtran.cls for IEEE Journals}

\IEEEtitleabstractindextext{%
\begin{abstract}
\justifying
Multimodal Large Language Models (MLLMs) have notably enhanced the performance of Multimodal Sentiment Analysis (MSA), yet their massive parameter scale leads to excessive resource consumption in training and inference, severely limiting model efficiency. To balance performance and efficiency for MSA, this paper innovatively proposes a novel Uncertainty-Aware Collaborative System (U-ACS) that integrates Uncertainty-aware Baseline Model (UBM) with MLLMs. U-ACS operates in three stages: First, all samples are processed by the UBM, retain high-confidence samples and forward low-confidence samples to the MLLM. Notably, to address the challenge that continuous outputs of regression tasks hinder uncertainty calculation, we innovatively convert the continuous sentiment label prediction task to a classification task, enabling a more accurate calculation of entropy and uncertainty. Second, the MLLM performs initial process. In this stage, high-confidence samples or low-confidence samples whose predictive sentiment polarity matches that of the UBM are deemed acceptable, while unqualified samples are forwarded for further processing. Finally, the MLLM performs secondary inference on remaining low-confidence samples using prompts augmented with prior rounds predictions as references. By aggregating results from the three stages, U-ACS preserves high MSA prediction accuracy while drastically boosting efficiency via offloading most simple samples to the UBM and minimizing MLLM processing volume. Extensive experiments verify that U-ACS maintains superior performance while significantly reducing computational overhead and resource consumption.

\end{abstract}

}

\maketitle

\IEEEdisplaynontitleabstractindextext

%
\IEEEpeerreviewmaketitle

\section{\textbf{Introduction}}\label{sec:Introduction}
\IEEEPARstart{I}n the real world, there exists a vast amount of multi-source and heterogeneous multimodal data. Such information carriers, which integrate different modalities including text, images, audio, and video, serve as the core medium for humans to perceive and cognize the entire objective world. Consequently, Multimodal Sentiment Analysis (MSA), which enables the comprehensive analysis of users’ emotional inclinations and opinion trends by leveraging the complementary information in multimodal data, has increasingly become a focal point of academic and industrial research \cite{Poria2017A,wang2020transmodality,mai2025learning}.

The landscape of multimodal learning \cite{8269806} has been profoundly reshaped by the emergence of powerful Multimodal Large Language Models (MLLMs) such as HumanOmni \cite{zhao2025humanomnilargevisionspeechlanguage} and VideoLLaMA2 \cite{damonlpsg2024videollama2}. Leveraging acoustic, textual, and visual information to predict human sentiment intensities and emotional states \cite{Poria2017A,liu2024ensemble}, MLLMs deliver enhanced performance in MSA. With their vast parameter counts and sophisticated reasoning capabilities, MLLMs demonstrates an unparalleled ability to interpret complex, nuanced human expressions. However, this remarkable accuracy comes at a significant cost that manifests as high inference latency, significant memory requirements, and considerable energy consumption, particularly problematic where resources are constrained or real-time responses are paramount.

In addition to MLLMs, academia and industry have also proposed a series of small-parameter specialized models \cite{wang2025dlf,mmsa} to address MSA tasks. Such lightweight models boast prominent advantages of high computational efficiency, fast inference speed, and low resource footprint, making them well-suited for deployment scenarios with limited computing power such as edge devices. However, their high efficiency is often achieved at the expense of overall inference performance: when confronted with samples featuring imbalanced data distribution, complex modality information, or missing data in partial modalities, the sentiment recognition accuracy and robustness of lightweight models will experience a significant decline, and their processing capability will be severely constrained.
Based on the inherent strengths and weaknesses of these two types of models, a key challenge has emerged in the MSA field: how to significantly improve the operational efficiency of models while maintaining their original inferential performance for MSA tasks? Therefore, exploring technical solutions that can integrate the advantages of both paradigms remains an important and open research issue in this domain.

\begin{figure}[t]
\centering
\includegraphics[width=0.95\linewidth]{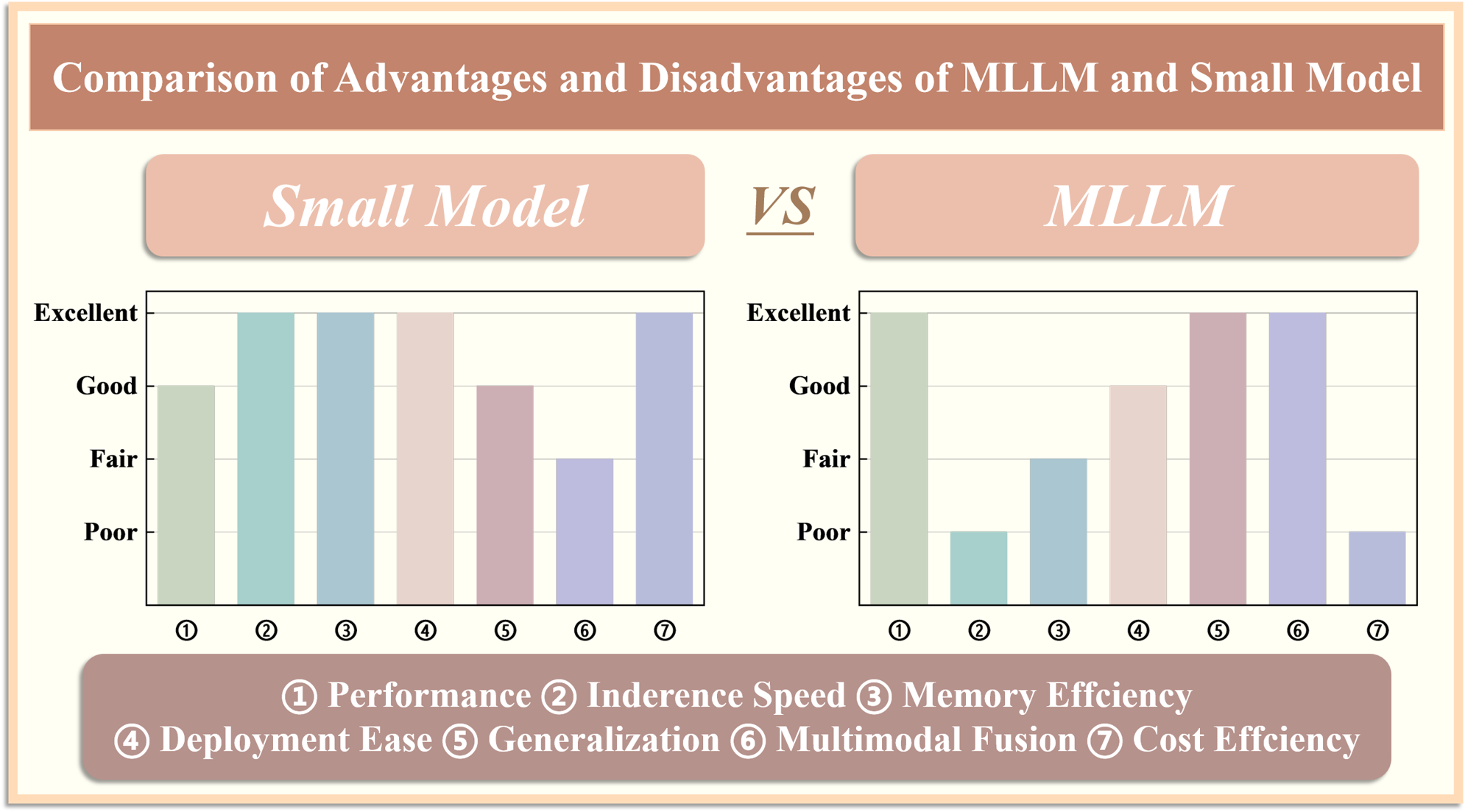}
\caption{Advantages and disadvantages of small models and MLLMs. Small models exhibit superior performance on efficiency, whereas large models demonstrate more advantages in performance.}
\label{fig:intro_concept}
\vspace{-0.3cm}
\end{figure}

Figure \ref{fig:intro_concept} systematically summarizes the core strengths and weaknesses of lightweight small models and MLLMs. It can be observed that there is a significant gap between the two types of models in terms of computational efficiency and inference speed: small models possess distinct lightweight advantages, whereas MLLMs exhibit high resource consumption. In terms of task performance, although MLLMs outperform small models overall, the sentiment analysis accuracy of small models still falls within an acceptable range. 
In view of the characteristics and shortcomings of these two types of models, this paper proposes a novel Uncertainty-Aware Collaborative System (U-ACS). This system innovatively achieves hierarchical collaboration between Uncertainty-aware Baseline Model (UBM) and MLLM, and realizes dynamic and adaptive allocation of computing resources between the two types of models through the design of a three-stage progressive processing workflow. 

Specifically, U-ACS quantifies uncertainty \cite{10287630,gao2024embracing} to evaluate the model’s predictive performance on specific samples. In the first stage of U-ACS, all samples are fed into the UBM for processing, with uncertainty scores computed for each sample. Samples with low uncertainty have high-confidence predictions from the UBM preserved; those with high uncertainty are forwarded to the MLLM for follow-up analysis. Notably, as regression tasks produce continuous outputs that hinder the calculation of corresponding uncertainty, we innovatively convert continuous sentiment label prediction task to a classification task, and build a three-class classifier to predict the discrete sentiment label (positive, negative and neutral). This operation allows us to accurately calculate sample entropy, which is then used to measure sample uncertainty.

Another two stages are operated within the MLLM module.
In the second stage, all samples with uncertainty exceeding the threshold in the UBM stage are fed into the MLLM for the first round of verification. Samples are categorized into three types according to their uncertainty scores and the consistency of sentiment polarity: high-confidence samples with low uncertainty, low uncertainty samples whose predicted sentiment polarities are consistent between UBM and MLLM, and high uncertainty samples with inconsistent sentiment polarities predicted by the two models. Among these, the prediction results of the first two types of samples can be directly adopted. The first type has high reliability due to sufficient confidence, while the second type, despite insufficient confidence, has consistent polarity judgments from the small and large models, making it unlikely that subsequent predictions will alter the core conclusion. Thus, only the third type of samples proceeds to the second round of in-depth inference in the MLLM. In the third stage, we embed the sentiment intensity predictions of these samples from the small model stage and the first-round MLLM results into the prompt as multi-dimensional reference bases, enabling the MLLM to make more accurate predictions for the remaining samples. Finally, by integrating the high-confidence results from the small model stage, the acceptable results from the first MLLM round, and the final discrimination results from the second MLLM round, we can obtain complete multimodal sentiment analysis results for all samples.

This hierarchical processing mechanism can significantly improve the operational efficiency of the model while ensuring the overall performance of MSA tasks, thereby effectively balancing the inherent trade-off between performance and efficiency. Our main contributions can be summarized as follows:
\begin{itemize}
  \item We propose a novel U-ACS that synthesizes MLLM with UBM to resolve the performance-versus-efficiency dilemma in MSA. To the best of our knowledge, this is the first small-large model collaboration scheme proposed for the MSA task.
  \item We design an uncertainty-guided cascade mechanism that intelligently routes samples, significantly reducing average computational cost by reserving the MLLM for only the most challenging cases. We also design multiple strategies to estimate predictive uncertainty for challenging regression tasks.
  \item Through extensive experiments, we demonstrate that our strategy achieves state-of-the-art performance on part of the key benchmarks across multiple MSA datasets, while also significantly improving the overall inference efficiency and reducing computational overhead substantially compared to MLLM.
\end{itemize}


\section{\textbf{Related Work}}\label{sec:Related Work}
\label{sec:format}
\subsection{Multimodal Sentiment Analysis}
The core task of MSA \cite{gandhi2023multimodal} is to fuse information from multimodalities such as text, images, audio, and video, so as to automatically identify and determine the sentiment tendency, intensity, or category implied in the target object (i.e., sentences, paragraphs, or video clips).
Existing studies have predominantly focused some research directions in multimodal learning: 
developing sophisticated fusion mechanisms to characterize cross-modal interactive relationships \cite{Poria2017A, Poria2017Convolutional_ori, huang2023cross, tsai-etal-2019-multimodal,TASLP,mai2022multimodal}; 
designing robust algorithmic frameworks to mitigate the adverse impacts of noisy modalities \cite{10008078, wang2023incomplete, Li_Yang_Lei_Wang_Wang_Su_Yang_Wang_Sun_Zhang_2024,mai2025supervised,10224356};
constructing weakly supervised and self-supervised learning paradigms to alleviate the constraint of scarce training data \cite{hycon}; 
and devising effective optimization algorithms to bridge inter-modality gaps while regularizing the learning process of unimodal data distributions \cite{hazarika2020misa, sun_meta, xiao2024neuroinspired}.
However, all the aforementioned core research endeavors are confronted with a critical challenge: the high complexity of MSA tasks necessitates the adoption of MLLMs to ensure satisfactory performance. Meanwhile, such task complexity also drives the design of sophisticated algorithms, which in turn leads to inefficient inference of MLLMs and substantial computational resource consumption.
To improve efficiency beyond traditional Transformers, State Space Models like MSAmba \cite{He_Liang_Peng_Xie_Khan_Song_Yu_2025} have demonstrated effective long-range modeling with lower computational complexity. However, these methods typically operate as monolithic systems, either prioritizing maximum performance or efficiency, without dynamic resource allocation mechanisms.

Meanwhile, Large Language Models (LLMs) have catalyzed significant advances in MSA \cite{yang2024large,mai2025injecting}. These methods often use lightweight adapters like MSE-Adapter \cite{yang2025mseadapterlightweightpluginendowing} to enable multimodal capabilities in frozen LLMs while reducing training costs. More recently, MLLMs have emerged, which can directly handle multimodal signals and project them into the feature space of LLMs and achieve better results \cite{zhao2025humanomnilargevisionspeechlanguage, wu2025enriching}.  While effective, these approaches invoke MLLMs for every sample, remaining computationally expensive for practical deployment.

\subsection{Collaboration of Large and Small Models}
An emerging trend across the machine learning field is the integration of large, high-capacity models and small, efficient models, which aims to optimize the trade-off between model performance and computational cost \cite{min2024synergetic,chen2025survey}. Such cascaded systems are designed to route straightforward inputs to lightweight models, with large models only invoked for handling challenging cases; this design has been validated effectively in low-latency computer vision and Natural Language Processing (NLP) tasks \cite{brown2020language}.

In the field of MSA, research on the collaboration mechanism between large and small models is still in the preliminary exploration stage, where existing methods mostly adopt a single large model or small model to complete tasks independently. In contrast, in other relevant machine learning fields, to facilitate the widespread deployment of models on resource-constrained devices, the collaboration mechanism between large and small models has gradually become a research hotspot jointly concerned by academia and industry. Chen et al. \cite{chen2025survey} categorized the collaboration patterns of large and small models into five major types: Pipeline Collaboration \cite{lv2025collaboration,zhang2024cogenesis,liu2025towards,chen2025knowledge,xuslmrec}, Auxiliary/Enhancement Collaboration \cite{shao2025division,huauxiliary,xu2025collab,deng2023mutual}, Knowledge Distillation-Driven Collaboration \cite{xu2024survey,xuslmrec,hendriks2025honey,guminillm}, and Integration/Fusion Collaboration \cite{naveed2025comprehensive,lv2025collaboration,subramanian2025small,donghymba,wanknowledge}, Hybrid/Routing Collaboration \cite{lv2025collaboration,wang2025mixllm,varangot2025doing}. Models characterized by distinct scales and performance capabilities are deployed on a task-dependent basis. A binary classifier is incorporated into HybridLLM \cite{yao2025toward} to assess query difficulty, with the system then directing queries to different models in accordance with the generated predictions.
The Collaborative Inference with Token-lEvel Routing (CITER) framework \cite{zhengciter}, a typical paradigm of routing collaboration, is adopted by this framework, with non-critical tokens being dispatched to the Small Language Model (SLM) for improved efficiency and critical tokens being transmitted to the LLM to ensure generation quality. Such fine-grained routing enables dynamic tuning of the efficiency-quality balance within the system.
Consistent with the above paradigm, our strategy is likewise classified as a Hybrid/Routing Collaboration approach.


Different from previous studies, our strategy prioritizes the improvement of model efficiency rather than the pursuit of pure performance gains. We aim to explore a viable paradigm for balancing efficiency and performance, and thus focus on developing an approach that can enhance efficiency while preserving the original performance level. To achieve this goal, this paper introduces a dedicated strategy that can fulfill this objective in MSA scenarios.

\section{\textbf{Model Architecture}}\label{sec:Algorithm}
\begin{figure*}[t]
\centering
\includegraphics[width=0.95\textwidth]{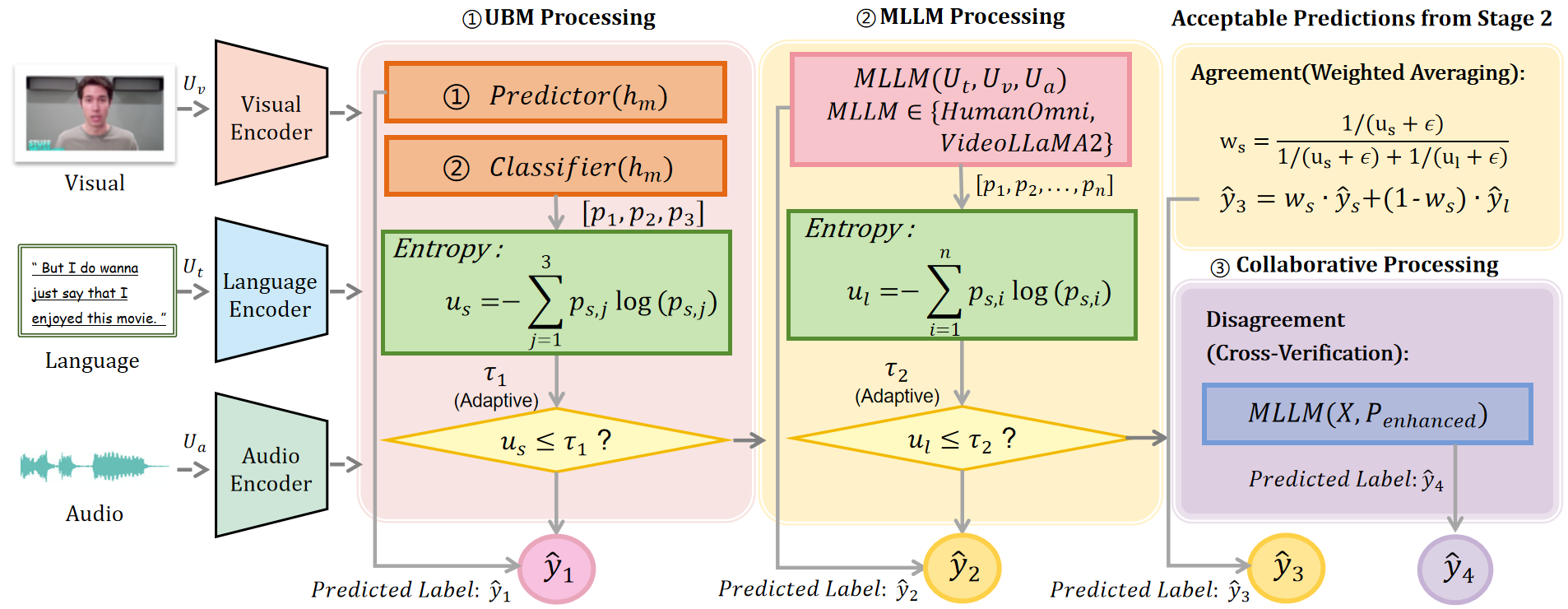}
\caption{The pipeline of the U-ACS. The diagram depicts the three-stage pipeline of U-ACS, where the first stage relies on the UBM, and the latter two stages are centered on the MLLM. The workflow is structured to leverage the complementary advantages of UBM and MLLM for enhanced MSA efficiency.}
\label{fig:framework_overview}
\vspace{-0.3cm}
\end{figure*}

In this section, we elaborate on the details of our proposed U-ACS framework, whose overall architecture is illustrated in Fig. \ref{fig:framework_overview}. The detailed algorithmic implementation of each component is presented in the subsequent subsections.

\subsection{Task Definition}
The goal of MSA is to predict the sentiment intensity from given multimodal inputs. In this task, we consider three different modalities, i.e., text ($t$), visual ($v$), and acoustic ($a$). The three modality is represented by sequences $\bm{U}_t \in \mathbb{R}^{N_t \times d_t}$, $\bm{U}_v \in \mathbb{R}^{N_v \times d_v}$, $\bm{U}_a \in \mathbb{R}^{N_a \times d_a}$,where $N_m$ denotes the sequence length, $d_m$ is the embedding dimension, and $m \in \{t, v, a\}$ indicates the specific modality type. The input feature sequence $\bm{U}_m \in \mathbb{R}^{N_m \times d_m}$ is mapping to unimodel representation $\bm{x}_m \in \mathbb{R}^{d_m}$ through the unimodel learning network $F^m$. The full structural details of unimodel learning network $F^m$ and fusion network $F^f$ in the subsequent subsections.

\subsection{Overall Architecture}
U-ACS establishes a synergistic paradigm that integrates UBM with MLLMs. Leveraging the complementary strengths of the two model types, where UBM excel in computational efficiency and MLLMs dominate in sentiment analysis performance, we design a tiered processing pipeline: UBM are tasked with predicting the sentiment intensity of easy samples, and difficult samples are forwarded to MLLMs for in-depth inference. This division of labor ensures that the framework maintains high sentiment analysis performance while significantly boosting overall inference efficiency. We will introduce the pipeline in UBM and MLLM separately.

\subsection{Pipeline of UBM} 
This subsection presents a comprehensive overview of the UBM, which is structured as follows: first, we decompose its core architecture, elaborate on the module functions and connection mechanisms of the encoder, fusion network, and predictor in sequence; second, we detail the uncertainty quantification method for the model’s outputs, which provides a theoretical basis for subsequent sample selection in the small-large model collaborative framework; finally, we present the mathematical formulation of the loss function adopted in the model training phase and offer corresponding interpretations.

\textbf{\textit{1) Model-specific Encoders}}

\textbf{Language Encoder:} For textual input processing, we adopt the BERT-base model \cite{bert} as the backbone encoder, where the English variant (bert-base-uncased) and Chinese variant (bert-base-chinese) are dynamically selected based on the language of the target dataset. Text sequences are first tokenized using the official BERT tokenizer, and the tokenized outputs are fed into the pre-trained BERT model to generate contextualized text representations:
\begin{equation}
\label{eq:text_encoder}
\begin{split}
\bm{H}_t &= \text{BERT}(\bm{U}_t) \in \mathbb{R}^{N_t \times d_{bert}}, \\
& \bm{x}_t = \bm{H}_t[0, :] \in \mathbb{R}^{d_{bert}}
\end{split}
\end{equation}
where $\bm{U}_t$ denotes the tokenized input of a single text sample, $\bm{H}_t$ represents the contextualized hidden state sequence output by BERT for the single text sample, with its dimension $\mathbb{R}^{N_t \times d_{\text{bert}}}$, $\bm{x}_t$ is the final text representation extracted from the 0-th token (i.e., the [CLS] token) of $\bm{H}_t$, which aggregates the global semantic information of the entire text sequence and has a dimension of $\mathbb{R}^{d_{\text{bert}}}$. Notably, the dimension $d_{\text{bert}} = 768$ is kept consistent across all datasets in our experiments for single-sample text representation.

\textbf{Audio Encoder:} Following Zadeh et al. \cite{zadeh-etal-2017-tensor} and Yu et al. \cite{yu-etal-2020-ch}, we ues pre-train ToolKits to extract the initial features in audio and visual modalities. For acoustic feature processing, following Self-MM \cite{mmsa}, we utilize a lightweight LSTM-based subnet to capture temporal dependencies, which has few parameters:
\begin{equation}
\label{eq:audio_encoder}
\begin{split}
\bm{H}_a = & \text{LSTM}(\bm{U}_a; \theta_a) \in \mathbb{R}^{N_a \times d_{a}}, \\
& \bm{x}_{a} = \bm{H}_{a}[-1, :] \in \mathbb{R}^{d_{a}}
\end{split}
\end{equation}
where $\bm{U}_a$ represents the input audio features, $\theta_a$ represents the full set of learnable parameters of the LSTM subnet, and $\bm{H}_a \in \mathbb{R}^{N_a \times d_a}$ is the end-state hidden vector output by the LSTM, which aggregates the global temporal dependencies of the entire audio sequence. $\bm{x}_a \in \mathbb{R}^{d_{a}}$ is the final audio representation with $d_{a} = 16$.

\textbf{Visual Encoder:} Similar to the audio encoder, we employ the same subnet architecture to process visual feature sequences:
\begin{equation}
\label{eq:visual_encoder}
\begin{split}
\bm{H}_v = & \text{LSTM}(\bm{U}_v; \theta_v) \in \mathbb{R}^{N_v \times d_{v}}, \\
& \bm{x}_{v} = \bm{H}_{v}[-1, :] \in \mathbb{R}^{d_{v}}
\end{split}
\end{equation}
where $\bm{U}_v$ represents the input audio features, $\theta_v$ represents the full set of learnable parameters of the LSTM subnet, and $\bm{H}_v \in \mathbb{R}^{N_v \times d_v}$ is the end-state hidden vector output by the LSTM, which aggregates the global temporal dependencies of the entire audio sequence. $\bm{x}_v \in \mathbb{R}^{d_{v}}$ is the final audio representation with $d_{v} = 32$.

\textbf{\textit{2) Multimodal Feature Fusion}}

The core function of the fusion network is to perform modal concatenation on the three types of unimodal representation, merging text, audio, and visual features into a holistic fused representation:
\begin{equation}
\label{eq:multimodal_fusion}
\begin{split}
& \bm{x}_{concat} = \bm{x}_t \Vert \bm{x}_a \Vert \bm{x}_v, \\
\bm{h}_{m} &= Fusion(\bm{x}_{concat}) \in \mathbb{R}^{d_{m}}
\end{split}
\end{equation}
where $\bm{x}_t$, $\bm{x}_a$, $\bm{x}_v$ represent unimodal representations (i.e., text, acoustic, visual), $\bm{x}_{concat}$ is the concatenated representation, $Fusion(\cdot)$ is a function that integrates the concatenated features into a fused representation $\bm{h}_m$ of dimension $d_{m}$. $W_1$ and $b_1$ are learnable parameters.

\textbf{\textit{3) Predictor}}

MSA tasks are generally treated as regression tasks, the sentiment predictor need to predict a sentiment label for each sample, and this label is a continuous value that reflects the fine-grained sentiment intensity of sample:
\begin{equation}
\label{eq:regression_head}
\hat{y}_s = W_r^{(2)} \text{ReLU}(W_r^{(1)} \bm{h}_m + b_r^{(1)}) + b_r^{(2)}
\end{equation}
where $\hat{y}_s$ is the predicted sentiment label for a single sample; $W_r^{(1)}$, $W_r^{(2)}$, $b_r^{(1)}$, $b_r^{(2)}$ are all learnable parameters of the regression predictor, with $\text{ReLU}$ acting as the activation function to capture non-linear relationships between multimodal features and sentiment scores. The fused multimodal representation $\bm{h}_m$ (in Eq. \ref{eq:multimodal_fusion}) is fed as input to this predictor.

\textbf{\textit{4) Uncertainty Estimation}}

A critical requirement of our UBM is to reliably estimate predictive uncertainty, which is essential for making effective routing decisions. As MSA is a regression task, estimating its predictive uncertainty is actually very difficult. Here we put forward three simple and reasonable uncertainty estimation approaches for MSA:\\
\begin{itemize}
    \item \textbf{Prediction-Truth Differences (PTD)}, which employs an additional predictor to measure the absolute difference between predicted and true labels, treating larger differences as indicators of higher uncertainty. However, predicting label differences is very difficult, requiring the model to have full discriminative power to estimate the labels of all samples. \\ 
    \item \textbf{Ensemble Variance (EV)}, which trains multiple diverse predictors on different data subsets and uses the variance among their predictions as an uncertainty measure. Although this approach is theoretically sound, it introduces computational overhead because each input must be processed by multiple predictors, conflicting with our efficiency goals.\\
    \item \textbf{Our Method – Classification-based Entropy (CBE).}
    To overcome the aforementioned limitations, we propose a classification-based entropy approach. Specifically, we transform the regression task into a three-class classification problem and employ a classifier to conduct the classification task, enabling the estimation of predictive uncertainty by computing the entropy of the output probability distribution. This method offers several key advantages: (i) computational efficiency achieved through a single forward pass; (ii) a principled, information-theoretic interpretation of uncertainty; and (iii) reducing the difficulty in uncertainty estimation, as the uncertainty of classification tasks is easier to estimate.
\end{itemize}

Specifically, alongside the primary regression predictor that outputs continuous sentiment label $\hat{y}_s$, we introduce an auxiliary classifier. The continuous sentiment scores are discretized into three categories according to their polarity:
\vspace{-0.3em}
\begin{equation}
\label{eq:sentiment_discretization}
y_{cls} = 
\begin{cases}
1 \; (Negative), & y < 0 \\
0 \; (Neutral),  & y = 0 \\
2 \; (Positive), & y > 0
\end{cases}
\end{equation}
The classification branch shares the same multimodal feature representation \(\bm{h}_{m}\) processed by the main regression predictor but employs a separate classifier:
\vspace{-0.3em}
\begin{equation}
\label{eq:classification_head}
\begin{split}
& \bm{z}_{cls} = Classifier(\bm{h}_{m}),\\
p_s =& Softmax(\bm{z}_{cls}) = [p_1, p_2, p_3]
\end{split}
\end{equation}
where ${z}_{cls}$ is a 3-dimensional vector derived from classifier processing, $p_s$ denotes the predicted probability distribution over the three sentiment classes.

During inference, uncertainty is quantified by the entropy of the predicted class distribution:
\vspace{-0.3em}
\begin{equation}
\label{eq:small_model_uncertainty_entropy}
u_s = H(p_s) = -\sum_{i=1}^{3} p_{s,i} \cdot \log(p_{s,i})
\end{equation}
where \(u_s\) denotes the uncertainty score, \(H(\cdot)\) is the entropy function, and \(p_{s,i}\) is the predicted probability of the \(i\)-th sentiment class. This entropy-based metric naturally reflects the model’s confidence: a high entropy value indicates greater uncertainty, corresponding to a more uniform distribution over classes, while a low entropy implies strong confidence concentrated on a single class \cite{xue2024ualign}.

The model is jointly optimized with a combined loss that integrates both regression and classification objectives:
\vspace{-0.3em}
\begin{equation}
\label{eq:combined_loss}
\mathcal{L}_{sup} = \mathcal{L}_{reg}(\hat{y}_s, y) + \mathcal{L}_{cls}(p_s, y_{cls})
\end{equation}
where \(\mathcal{L}_{reg}\) represents the Mean Squared Error (MSE) for regression, \(\mathcal{L}_{cls}\) is the Cross-Entropy (CE) loss for classification, and \(y_{cls}\) is the discretized ground truth label.

\subsection{Pipeline of MLLM}
We adopt HumanOmni \cite{yang2025mseadapterlightweightpluginendowing} and VideoLLaMA2 \cite{damonlpsg2024videollama2} as expert oracles to handle high-uncertainty and difficult samples in the MSA task. HumanOmni serves as the core MLLM, whereas VideoLLaMA2 is leveraged for validating the generalization capability of our approach. In the following, we will elaborate on the specific deployment schemes and application scenarios in this study.

The applied HumanOmni and VideoLLaMA2 are both 7-billion-parameter MLLM. These models’ output are formulated as:
\vspace{-0.3em}
\begin{equation}
\label{eq:large_model_output}
\begin{split}
& \mathrm{logits}_l = \text{MLLM}(\bm{U}_t, \bm{U}_a, \bm{U}_v), \\
\text{MLLM} & \in \{\text{HumanOmni}, \text{VideoLLaMA2}\}
\end{split}
\end{equation}
where \(\bm{U}_t, \bm{U}_a, \bm{U}_v\) denote the input features from text, audio, and visual modalities, respectively. $\mathrm{logits}_l$ is the original probability for samples. Based on the obtained raw probability distribution, we can not only determine the label with the highest prediction likelihood for the corresponding sample, but also derive the probability vector from this distribution to further calculate the uncertainty of each sample:
\vspace{-0.3em}
\begin{equation}
\begin{split}
\label{eq:large_model_prob}
\hat{y}_l = \underset{y_k \in y_l}{\text{argmax}} \ \text{logits}_l(y_k), & \\
p_l = \mathrm{Softmax}(\mathrm{logits}_l) = [p_1, p_2, & \ldots, p_n]
\end{split}
\end{equation}
where $\hat{y}_l$ is the predicted sentiment score, $y_l$ is the scope of sentiment score, ${p}_l$ represents the predicted probability vector which is come from $\mathrm{logits}_l$, the definition of $\mathrm{logits}_l$ can be seen in Eq. \ref{eq:large_model_output}. \(p_i\) is the probability of the \(i\)-th class, and \(n\) is the total number of sentiment classes. We quantify uncertainty by calculating the entropy of this distribution:

\vspace{-0.3em}
\begin{equation}
\label{eq:large_model_uncertainty}
u_l = H({p}_l) = -\sum_{i=1}^n p_{l,i} \cdot \log(p_{l,i}),
\end{equation}
where \(u_l\) denotes the uncertainty score and \(H(\cdot)\) is the entropy function. Higher entropy corresponds to higher uncertainty, providing a principled confidence measure for the model's predictions. As output consists of multiple tokens, we average the uncertainties of all label-related output tokens to obtain the final uncertainty. The method for determining the uncertainty threshold is explicated in Subsection~\ref{subsec:uncertainty_threshold}.

To efficiently adapt HumanOmni and VideoLLaMA2 to our task, we apply LoRA \cite{hu2022lora}, which decomposes the weight update into low-rank matrices \(A \in \mathbb{R}^{d \times r}\) and \(B \in \mathbb{R}^{r \times d}\), with rank \(r\) much smaller than \(d\). The adapted weight matrix is given by \(W_{new} = W + A \times B\), where \(W\) is the original weight matrix and only \(A\) and \(B\) are trainable during fine-tuning.

\subsection{Pipeline of U-ACS}
Building upon the uncertainty thresholds \(\tau_1\) and \(\tau_2\) (see Section~\ref{subsec:uncertainty_threshold}), we now present the overall inference workflow, which is shown in Algorithm~\ref{alg:algorithm}. This workflow employs a dynamic collaboration strategy that sequentially leverages UBM and MLLM, guided by their uncertainty estimates, to efficiently and accurately produce the final prediction. The process proceeds through three stages:

\textbf{Stage 1: UBM processing.}  
Compute the baseline prediction \(\hat{y}_s\) and uncertainty \(u_s\). If the uncertainty is below the adaptive validation-calibrated threshold \(\tau_1\), accept this prediction directly. Otherwise, proceed to the MLLM Refinement Stage:
\begin{equation}
\label{eq:stage1_decision}
\hat{y}_1 = 
\begin{cases}
\hat{y}_s, & u_s \leq \tau_1 \\
\text{Proceed to Stage 2}, & u_s > \tau_1
\end{cases}
\end{equation}

\textbf{Stage 2: MLLM processing.}  
For samples with higher uncertainty, invoke the MLLM to obtain prediction \(\hat{y}_l\) along with its uncertainty \(u_l\). Accept the prediction if the uncertainty is below the second threshold \(\tau_2\):
\vspace{-0.3em}
\begin{equation}
\label{eq:stage2_decision}
\hat{y}_2 = 
\begin{cases}
\hat{y}_l, & u_l \leq \tau_2 \\
\text{Proceed to Stage 3}, & u_l > \tau_2
\end{cases}
\end{equation}

For cases still deemed highly uncertain, apply collaborative strategies based on whether two models agree on sentiment polarity:
\begin{itemize}
    \item \emph{Agreement (Weighted Averaging):}  
    If the predictions of both models share the same sentiment polarity, compute adaptive weights based on uncertainty and combine their outputs:
    \vspace{-0.3em}
    \begin{equation}
    \label{eq:weight_avg}
    \begin{split}
    w_s &= \frac{1/(u_s+\epsilon)}{1/(u_s+\epsilon) + 1/(u_l+\epsilon)}, \\
    \hat{y}_3 &= w_s \cdot \hat{y}_s + (1 - w_s) \cdot \hat{y}_l
    \end{split}
    \end{equation}

    \item \emph{Disagreement (Cross-Verification):}  
    If two models disagree, augment the prompt with previous predictions and uncertainties, and re-predict through the Stage 3.
\end{itemize}

\textbf{Stage 3: Collaborative Processsing.}
Low-confidence samples generated in Stage 2 are sent to Stage 3 for re-prediction, with the prior prediction outputs of Stage 1 and Stage 2 embedded into the prompt as references:
\vspace{-0.3em}
\begin{equation}
\label{eq:cross_verify}
\begin{split}
& \hat{y}_4 = \text{MLLM}(X, P_{enhanced}), \\
\text{MLLM} & \in \{\text{HumanOmni}, \text{VideoLLaMA2}\}
\end{split}
\end{equation}

where \(P_{enhanced}\) provides richer context by including UBM’s prediction \(\hat{y}_s\) and uncertainty \(u_s\), along with the prediction $\hat{y}_l$ and uncertainty  \(u_l\) of the MLLM. $\epsilon$ denotes a small constant parameter with a fixed value of $10^{-12}$. 

Finally, we concatenate the output results of three stages to obtain the prediction results for all samples, and calculate the accuracy of our algorithm based on these prediction results:
\begin{equation}
\hat{y} = \hat{y}_1 \Vert \hat{y}_2 \Vert \hat{y}_3 \Vert \hat{y}_4
\end{equation}
where $\hat{y}$ denotes the prediction results for all samples, and $\hat{y}_1$ represents the prediction result of Stage 1, $\hat{y}_2$ and $\hat{y}_3$ those of Stage 2, and $\hat{y}_4$ that of Stage 3.

\subsection{Setting Criterion for Uncertainty Threshold}
\label{subsec:uncertainty_threshold}
We adopt a principled statistical approach to determine the uncertainty thresholds for both the UBM and the MLLM. This method analyzes the uncertainty distributions on the validation set, providing a theoretically grounded yet computationally efficient way to select thresholds. 

The threshold selection procedure is applied consistently to both models. First, we determine the threshold value by calculating the mean of uncertainty values across all samples. The threshold derived from this method has yielded favorable performance in our experiments. To further optimize the results, we finally incorporate the Gaussian fitting strategy and apply this enhanced method to compute the specific thresholds corresponding to each of the two models. For each model, we partition the validation samples into two groups based on whether the model's predicted polarity matches the ground truth. We define \(\mathbb{D}_{same}\) as the set of samples where the prediction and ground truth share the same polarity (both positive or both negative), and \(\mathbb{D}_{opposite}\) as the set where they differ in polarity:
\vspace{-0.3em}
\begin{equation}
\label{eq:threshold_stats_mean}
\begin{split}
\mu_{same} &= \mathbb{E}_i[(u)_i \mid i \in \mathbb{D}_{same}], \\
\mu_{opposite} &= \mathbb{E}_i[ (u)_i \mid i \in \mathbb{D}_{opposite}]
\end{split}
\end{equation}
where \((u)_i\) denotes the uncertainty for sample $i$.
Finally, we apply this method with Gaussian fitting to compute the specific thresholds for both models:
\vspace{-0.3em}
\begin{equation}
\label{eq:model_threshold}
\begin{split}
\tau_1 &= {(1-\lambda)\cdot\mu_{same}^{(s)}} + \lambda\cdot\mu_{opposite}^{(s)}+\beta, \\
\tau_2 &= {(1-\lambda)\cdot\mu_{same}^{(l)}} + \lambda\cdot\mu_{opposite}^{(l)}+\beta,
\end{split} 
\end{equation}
where \(\tau_1\) corresponds to UBM and \(\tau_2\) to the MLLM. The superscripts \((s)\) and \((l)\) denote statistics computed from UBM and MLLM predictions, respectively. $\lambda$ is a hyperparameter for weight with a default value of 0.5, while $\beta$ is a hyperparameter for bias with a default value of 0.0.

\begin{algorithm}[tb]
\caption{Uncertainty-Aware Collaborative Inference}
\label{alg:algorithm}
\textbf{Input}: Sample \(X = (\bm{U}_t, \bm{U}_a, \bm{U}_v)\), UBM \(M_s\), MLLM \(M_l\), Uncertainty Thresholds \(\tau_1, \tau_2\); \\
\textbf{Output}: Final prediction \(\hat{y}\);
\begin{algorithmic}[1]
\STATE \(\hat{y}_s, p_s \leftarrow M_s(X);\)
\STATE \(u_s \leftarrow -\sum_{i=1}^{3} p_{s,i} \cdot \log (p_{s,i});\)
\IF{\(u_s \leq \tau_1\)}
    \STATE \(\hat{y}_1 \leftarrow \hat{y}_s;\)
\ELSE
    \STATE \(\hat{y}_l, p_l \leftarrow M_l(X);\)
    \STATE \(u_l \leftarrow -\sum_{i=1}^n p_{l,i} \cdot \log(p_{l,i});\)
    \IF{\(u_l \leq \tau_2\)}
        \STATE \(\hat{y}_2 \leftarrow \hat{y}_l;\)
    \ELSE
        \IF{\(\hat{y}_s\) and \(\hat{y}_l\) are the same sentiment polarity}
            \STATE \(w_s \leftarrow \frac{1/(u_s+\epsilon)}{1/(u_s+\epsilon) + 1/(u_l+\epsilon)};\)
            \STATE \(\hat{y}_3 \leftarrow w_s \cdot \hat{y}_s + (1 - w_s) \cdot \hat{y}_l;\)
        \ELSE
            \STATE Construct \(P_{enhanced}\) from \(\hat{y}_s, u_s, \hat{y}_l, u_l;\)
            \STATE \(\hat{y}_4 \leftarrow M_l(X, P_{enhanced});\)
        \ENDIF
    \ENDIF
\ENDIF
\RETURN \(\hat{y} = \hat{y}_1 \Vert \hat{y}_2 \Vert \hat{y}_3 \Vert \hat{y}_4\)
\end{algorithmic}
\end{algorithm}

\section{\textbf{Experiments}}\label{sec:Experiments}
In this section, we evaluate our strategy on three different multimodal datasets on MSA. We present experiment details and results as follows.

\subsection{Datasets}

\subsubsection{\textbf{CMU-MOSI}}
CMU-MOSI (MOSI) \cite{Zadeh2016Multimodal} contains 2,199 video segments that study the interaction patterns between facial gestures and spoken words for sentiment prediction. The dataset provides fine-grained sentiment annotations on a continuous scale from strongly negative to strongly positive, making it ideal for evaluating both regression and classification capabilities of multimodal models.

\subsubsection{\textbf{CMU-MOSEI}}
CMU-MOSEI (MOSEI) \cite{zadeh2018multimodal} is a larger multimodal dataset for sentiment analysis and emotion recognition, containing 23,453 annotated video segments from 1,000 distinct speakers across 250 topics, serving as the next generation expansion of MOSI. The dataset's scale and diversity provide a robust evaluation platform for assessing model generalization across different speakers and topics.

\subsubsection{\textbf{CH-SIMS v1}} CH-SIMS v1 (SIMS) \cite{yu-etal-2020-ch} is a Chinese multimodal dataset which contains 2,281 refined video segments in the wild with both multimodal and independent unimodal annotation. This dataset enables cross-lingual validation of our approach and demonstrates the framework's robustness across different languages and cultural contexts.

\subsection{Evaluation Metrics}
Following standard MSA evaluation protocols, we report comprehensive performance metrics across different granularities:

\subsubsection{Classification Metrics}
\begin{itemize}
    \item \textbf{7-class accuracy (Acc7):} Accuracy for fine-grained sentiment classification across seven sentiment levels, evaluating the model's ability to distinguish subtle sentiment differences.
    \item \textbf{Binary accuracy (Acc2):} Accuracy for positive/negative classification, which is crucial for practical applications requiring simple sentiment polarity detection.
    \item \textbf{F1-score:} Harmonic mean of precision and recall for binary classification, providing a balanced measure that accounts for both false positives and false negatives.
    \item \textbf{5-class accuracy (Acc5):} Used specifically for CH-SIMS dataset, evaluating performance on five sentiment categories.
    \item \textbf{3-class accuracy (Acc3):} Used for CH-SIMS dataset, assessing performance on positive, negative, and neutral sentiment categories.
\end{itemize}

\subsubsection{Regression Metrics}
\begin{itemize}
    \item \textbf{Mean Absolute Error (MAE):} Measures the average absolute difference between predicted and ground truth sentiment scores, providing insight into prediction precision.
    \item \textbf{Pearson correlation (Corr):} Evaluates the linear correlation between predictions and ground truth, indicating how well the model captures sentiment intensity relationships.
\end{itemize}

\subsubsection{Efficiency Metrics}
\begin{itemize}
    \item \textbf{Inference Time:} Runtime measured in seconds to assess computational efficiency.
\end{itemize}

\subsection{Baseline and Compared Models}
This section describes the baseline models used in our experiments, including MISA, HumanOmni, VideoLLaMA2 and so on.

\subsubsection{MISA \cite{hazarika2020misa}}
Learns modality-invariant (shared) and modality-specific (unique) representations from multimodal data.

\subsubsection{Low-rank Multimodal Fusion (LMF) \cite{liu-etal-2018-efficient-low}}
Employs low-rank tensors for efficient multimodal fusion, significantly reducing computational complexity compared to traditional tensor-based methods.

\subsubsection{Multimodal Transformer (MulT) \cite{tsai-etal-2019-multimodal}}
A Multimodal Transformer designed to handle unaligned multimodal sequences through the use of directional pairwise cross-modal transformers.

\subsubsection{Multimodal Adaptation Gate-BERT (MAG-BERT) \cite{rahman-etal-2020-integrating}}
Integrates non-verbal modalities (audio and visual) into pre-trained language models like BERT using a multimodal attention gating mechanism.

\subsubsection{Self-Supervised Multi-task Multimodal sentiment analysis network (Self-MM) \cite{mmsa}}
A self-supervised framework that automatically generates unimodal labels to guide a multi-task learning process, avoiding the need for manual unimodal annotation.

\subsubsection{Contrastive FEature DEcomposition framework (ConFEDE) \cite{yang-etal-2023-confede}}
Decomposes features from each modality into similarity and dissimilarity components using a unified contrastive learning framework.


\subsubsection{Disentangled-Language-Focused (DLF)  multimodal representation learning framework \cite{wang2025dlf}}
A framework that prioritizes the language modality by disentangling shared and specific features and using a Language-Focused Attractor to leverage complementary information.

\subsubsection{Tensor Fusion Network (TFN) \cite{zadeh-etal-2017-tensor}}
A pioneering model that uses tensor fusion via outer products to capture high-order, complex inter-modal dynamics.

\subsubsection{Knowledge-Guided Dynamic Modality Attention Fusion Framework (KuDA) \cite{feng-etal-2024-knowledge}}
Dynamically selects the dominant modality for a given sample using sentiment knowledge, rather than statically assuming text is always the most important.

\subsubsection{VideoLLaMA2 \cite{damonlpsg2024videollama2}}
A powerful MLLM with strong spatial-temporal and audio understanding capabilities. It is used as a standalone baseline to benchmark the efficiency and performance of our U-ACS framework.

\subsubsection{HumanOmni \cite{zhao2025humanomnilargevisionspeechlanguage}}
A 7-billion-parameter MLLM that serves as the expert oracle within our U-ACS framework. It is LoRA-adapted to process high-uncertainty samples that are escalated from the baseline model.

\subsection{Experimental Details}

We develop our approach using PyTorch on NVIDIA GeForce RTX 4090 with CUDA 13.0 and torch 2.2.1. 

UBM is trained using the PyTorch Framework. The model employs component-specific learning rates and weight decay values that vary across datasets. Table~\ref{tab:optimization_strategy} presents the detailed optimization parameters for each dataset.

For HumanOmni and VideoLLaMA2, we configure the training epochs, learning rate, LoRA Rank, and  LoRA $\alpha$ to 1, 2e-5, 128, and 256 respectively.

\begin{table}[t]
\caption{Component-specific optimization configurations.}
\centering
\small
\setlength{\tabcolsep}{6pt}
\resizebox{0.45\textwidth}{!}{
\begin{tabular}{l|c|ccc}
\hline
\textbf{Component} & \textbf{Parameter} & \textbf{MOSI} & \textbf{MOSEI} & \textbf{SIMS} \\
\hline
\multirow{2}{*}{BERT} & Learning Rate & $5 \times 10^{-5}$ & $5 \times 10^{-5}$ & $5 \times 10^{-5}$ \\
                      & Weight Decay & $1 \times 10^{-3}$ & $1 \times 10^{-5}$ & $1 \times 10^{-3}$ \\
\hline
\multirow{2}{*}{Audio LSTM} & Learning Rate & $1 \times 10^{-3}$ & $5 \times 10^{-3}$ & $5 \times 10^{-3}$ \\
                            & Weight Decay & $1 \times 10^{-3}$ & $0$ & $1 \times 10^{-2}$ \\
\hline
\multirow{2}{*}{Video LSTM} & Learning Rate & $1 \times 10^{-4}$ & $1 \times 10^{-3}$ & $5 \times 10^{-3}$ \\
                            & Weight Decay & $1 \times 10^{-3}$ & $0$ & $1 \times 10^{-2}$ \\
\hline
\multirow{2}{*}{Others} & Learning Rate & $5 \times 10^{-4}$ & $1 \times 10^{-3}$ & $1 \times 10^{-3}$ \\
                        & Weight Decay & $1 \times 10^{-3}$ & $1 \times 10^{-2}$ & $1 \times 10^{-3}$ \\
\hline
\end{tabular}}
\label{tab:optimization_strategy}
\end{table}

\subsection{Result on Multimodal Sentiment Analysis}
To assess the performance of U-ACS, we utilized three well-recognized MSA datasets. HumanOmni is employed as the core MLLM in U-ACS.
As presented in Table \ref{tab:main_results}, U-ACS demonstrates superior performance, achieving state-of-the-art results in some evaluation metrics while maintaining computational efficiency. 
Compared to HumanOmni, U-ACS achieves 0.5 points improvement in Acc2 and F1 score on MOSI, and achieves 0.3 points improvement in Acc2, 0.4 points improvement in F1 score on MOSEI. Our approach achieves superior performance over conventional multimodal methods on all metrics.
Table \ref{tab:chsims_results} presents the robustness validation results of U-ACS in cross-lingual scenarios. On SIMS, the model maintains leading performance across most metrics, including Acc3, Acc2 and F1-score.
Furthermore, U-ACS significantly outperforms all traditional multimodal methods, demonstrating the effectiveness of uncertainty-guided collaboration across different languages and data characteristics.

It is noteworthy that while U-ACS consistently improves binary classification metrics, the standalone MLLM occasionally retain an edge in regression-focused metrics like MAE and Corr. This outcome is not an unintended side effect but a direct consequence of our framework's dual design philosophy: to maximize both classification accuracy and computational efficiency. Our core uncertainty estimation and sample routing mechanisms are designed precisely to serve these two goals, which also explains the trade-off in regression performance.
Specifically, the UBM estimates uncertainty by converting the regression task into a simpler three-class classification problem and calculating prediction entropy. Furthermore, the routing thresholds are determined by evaluating whether the predicted polarity aligns with the ground truth, a process that is both fast and directly tied to the primary goal of Acc2. The entire cascade is therefore architected to quickly filter the majority of `easy' samples using a cheap model, drastically reducing the average inference time and reserving the expensive MLLM only for challenging cases.
This design creates a deliberate trade-off. The standalone MLLM, which processes every sample, is singularly focused on minimizing the error of fine-grained sentiment scores across the board, thus naturally excelling at metrics like MAE and Corr that measure the precision of continuous values. Our U-ACS, in contrast, prioritizes rapidly and correctly identifying the sentiment polarity for the bulk of the data. It will accept a less precise regression value from the small model if the polarity confidence is high, because doing so is far more efficient. Consequently, we achieve superior performance in Acc2 and F1-scores and a significant reduction in runtime, at the minor, and acceptable, cost of precision in sentiment intensity prediction for some samples.

\begin{table*}[t]
\caption{Performance comparison on MOSI and MOSEI datasets. Best results are in \textbf{bold}, second-best are \underline{underlined}. $^{\dagger}$ means the results are from \cite{feng-etal-2024-knowledge}.}
\centering
\small
\vspace{-0.1cm}
\begin{tabular}{l|ccccc|ccccc}
\hline
\multirow{2}{*}{Method} & \multicolumn{5}{c|}{MOSI} & \multicolumn{5}{c}{MOSEI} \\
& Acc7$\uparrow$ & Acc2$\uparrow$ & F1$\uparrow$ & MAE$\downarrow$ & Corr$\uparrow$ & Acc7$\uparrow$ & Acc2$\uparrow$ & F1$\uparrow$ & MAE$\downarrow$ & Corr$\uparrow$ \\
\hline
MISA \cite{hazarika2020misa} & 42.3 & 83.4 & 83.6 & 0.783 & 0.761 & 52.2 & 85.5 & 85.3 & 0.555 & 0.756 \\
LMF$^{\dagger}$ \cite{liu-etal-2018-efficient-low} & 33.8 & 79.2 & 79.2 & 0.950 & 0.651 & 51.6 & 83.5 & 83.4 & 0.575 & 0.716 \\
MulT$^{\dagger}$ \cite{tsai-etal-2019-multimodal} & 36.9 & 79.7 & 79.6 & 0.879 & 0.702 & 52.8 & 84.6 & 84.5 & 0.559 & 0.733 \\
MAG-BERT \cite{rahman-etal-2020-integrating} & 43.6 & 84.4 & 84.6 & 0.727 & 0.781 & 52.7 & 84.8 & 84.7 & 0.543 & 0.755 \\
Self-MM \cite{mmsa} &  45.8 &   84.9  &   84.8  & 0.731  & 0.785  &  53.0 &   85.2  &   85.2  &  0.540 & 0.763 \\
ConFEDE \cite{yang-etal-2023-confede} & 42.3 & 85.5 & 85.5 & 0.742 & 0.784 & 54.9 & 85.8 & \textbf{85.8} & 0.522 & 0.780 \\
DLF \cite{wang2025dlf}  & 47.1  & 85.1  &  85.0  &  0.731 & 0.781  & 53.9  & 85.4  &  85.3 & 0.536 & 0.764 \\
HumanOmni \cite{zhao2025humanomnilargevisionspeechlanguage} & \textbf{52.8} & \underline{91.3} & \underline{91.3} & \textbf{0.549} & \textbf{0.881} & \textbf{58.6} & \underline{86.1} & \underline{85.4} & \textbf{0.483} & \textbf{0.807} \\
\hline
\textbf{U-ACS} & \underline{51.2} & \textbf{91.8} & \textbf{91.8} & \underline{0.585} & \underline{0.868} & \underline{55.4} & \textbf{86.4} & \textbf{85.8} & \underline{0.506} & \underline{0.784} \\
\hline
\end{tabular}
\label{tab:main_results}
\end{table*}

\begin{table}[t]
\caption{Performance comparison on SIMS dataset. Best results are in \textbf{bold}, second-best are \underline{underlined}. $^{\dagger}$ means the results are from \cite{feng-etal-2024-knowledge}. }
\centering
\small
\setlength{\tabcolsep}{4pt}
\begin{tabular}{l|ccccc}
\hline
\multirow{2}{*}{Method} & \multicolumn{5}{c}{SIMS} \\
& Acc3$\uparrow$ & Acc2$\uparrow$ & F1$\uparrow$ & MAE$\downarrow$ & Corr$\uparrow$ \\
\hline
MISA \cite{hazarika2020misa} & - & 76.5 & 76.6 & 0.447 & 0.563 \\
TFN$^{\dagger}$ \cite{zadeh-etal-2017-tensor} & 65.1 & 78.4 & 78.6 & 0.432 & 0.591  \\
MulT$^{\dagger}$ \cite{tsai-etal-2019-multimodal} & 64.8 & 78.6 & 79.7 & 0.453 & 0.564 \\
MAG-BERT \cite{rahman-etal-2020-integrating} & - & 74.4 & 71.8 & 0.492 & 0.399 \\
DEVA \cite{wu2025enriching} & 65.4 & 79.6 & 80.3 & 0.424 & 0.583 \\
KuDA$^{\dagger}$\cite{feng-etal-2024-knowledge} & 66.5 & 80.7 & 80.7 & 0.408 & 0.613\\
HumanOmni \cite{zhao2025humanomnilargevisionspeechlanguage}& \underline{72.9} & \underline{85.1} & \underline{85.0} & \textbf{0.327} & \textbf{0.749} \\
\hline
\textbf{U-ACS} & \textbf{73.1} & \textbf{85.8} & \textbf{85.1} & \underline{0.354} & \underline{0.691} \\
\hline
\end{tabular}
\label{tab:chsims_results}
\end{table}

\subsection{Study on Computational Efficiency}
\label{sec:efficiency}
In this subsection, we analyze the computational efficiency of HumanOmni. Table \ref{tab:humanomni_efficiency} compares the computational costs, Acc2, and MAE of different approaches.

Compared with HumanOmni, our strategy achieves the optimal performance in the Acc2 metric on MOSI, MOSEI, and SIMS datasets, while attaining significant efficiency improvements.  
Specifically, on the MOSI dataset, the runtime of U-ACS is only 33\% of the time HumanOmni takes, which significantly reduces the inference time while maintaining overall performance.

These results indicate that the proposed strategy, while ensuring that the model performance reaches the level of MLLM, has significantly shortened inference time and reduced model complexity, greatly improving operational efficiency and ultimately achieving a balance between performance and efficiency. We will further analyze the performance and adaptability of the proposed strategy across different MLLMs in Section \ref{sec:MLLM}.

\begin{table}[t]
\caption{Computational efficiency comparison on HumanOmni. Time is measured in seconds per sample. Best results are in \textbf{bold}, second-best are \underline{underlined}.}
\centering
\small
\setlength{\tabcolsep}{5pt}
\begin{tabular}{l|l|ccc}
\hline
Dataset & Method & Runtime$\downarrow$ & Acc2$\uparrow$ & MAE$\downarrow$ \\
\hline
\multirow{3}{*}{MOSI} 
& UBM & \textbf{15.35} & 86.1 & 0.716 \\
& HumanOmni \cite{zhao2025humanomnilargevisionspeechlanguage} & 379.18 & \underline{91.3} & \textbf{0.549} \\
& \textbf{U-ACS} & \underline{125.76} & \textbf{91.8} & \underline{0.585} \\
\hline
\multirow{3}{*}{MOSEI} 
& UBM & \textbf{35.02} & \underline{86.3} & 0.527 \\
& HumanOmni \cite{zhao2025humanomnilargevisionspeechlanguage} & 9815.56 & 86.1 & \textbf{0.483} \\
& \textbf{U-ACS} & \underline{5561.18} & \textbf{86.4} & \underline{0.506} \\
\hline
\multirow{3}{*}{SIMS} 
& UBM & \textbf{18.76} & 80.5 & 0.409 \\
& HumanOmni \cite{zhao2025humanomnilargevisionspeechlanguage} & 470.61 & \underline{85.1} & \textbf{0.327} \\
& \textbf{U-ACS} & \underline{260.58} & \textbf{85.8} & \underline{0.354} \\
\hline
\end{tabular}
\label{tab:humanomni_efficiency}
\end{table}

\subsection{Study and Analysis on Different MLLMs}
\label{sec:MLLM}
To verify the generalization ability and robustness of the proposed U-ACS, we conduct experiments by integrating it with a different powerful MLLM: VideoLLaMA2. This subsection aims to demonstrate that the U-ACS strategy is not reliant on a specific model architecture but can effectively enhance the performance of different MLLM.

As shown in Table \ref{tab:generalization_MLLM}, the U-ACS consistently improves Acc2 and F1-score for VideoLLaMA2 across the MOSI and SIMS datasets. For instance, when paired with VideoLLaMA2, U-ACS boosts the Acc2 score by 2.2 points on MOSI (from 88.3\% to 90.5\%) and a remarkable 4.6 points on SIMS (from 81.0\% to 85.6\%). A similar trend is observed with HumanOmni, where U-ACS improves the Acc2 by 0.5 points on MOSI and 0.7 points on SIMS.
Notably, our method yields optimal Acc2 performance across all three datasets for VideoLLaMA2, which aligns with the findings for HumanOmni.
While the standalone MLLMs occasionally perform better on other metrics such as MAE or Corr, the significant and consistent improvement in Acc2 and F1 highlights the effectiveness of our collaborative system. This outcome aligns with our design, which optimizes for binary polarity classification by using coarse-grained positive and negative categories for thresholding and uncertainty estimation. Moreover, in the UBM, we transform continuous sentiment labels into positive, negative, and neutral classes. These designs make the optimization of Acc2 (positive versus negative) more targeted, thereby highlighting significant advantages. 

Meanwhile, the proposed strategy not only achieves optimal performance across both Acc2 and MAE on the SIMS dataset, but also reduces the computational time by nearly 50\% on both the MOSI and SIMS datasets in Table \ref{tab:videollama2_efficiency}.
These results confirm that U-ACS is a model-agnostic framework that successfully leverages MLLMs to handle difficult samples, validating the broad applicability and effectiveness of our approach.

\begin{table}[t]
\caption{Generalization Evaluation on Different MLLMs. Best results are in \textbf{bold}, second-best are \underline{underlined}.}
\centering
\small
\setlength{\tabcolsep}{4pt}
\resizebox{0.47\textwidth}{!}{
\begin{tabular}{l|l|ccccc}
\hline
Dataset & Description & Acc7$\uparrow$  & Acc2$\uparrow$  & F1$\uparrow$  & MAE$\downarrow$  & Corr$\uparrow$  \\
\hline
\multirow{2}{*}{MOSI} 
& VideoLLaMA2 \cite{damonlpsg2024videollama2} & \textbf{49.8} & \underline{88.3} & \underline{88.3} & \textbf{0.575} & \textbf{0.875} \\
& \textbf{U-ACS} & \underline{48.9} & \textbf{90.5} & \textbf{90.5} & \underline{0.625} & \underline{0.847}  \\
\cline{2-7}
\hline
\multirow{2}{*}{SIMS} 
& VideoLLaMA2 \cite{damonlpsg2024videollama2}  & \underline{74.0} & \underline{81.0} & \underline{81.7} & \underline{0.400} & \textbf{0.718} \\
& \textbf{U-ACS} & \textbf{74.4} & \textbf{85.6} & \textbf{85.6} & \textbf{0.392} & \underline{0.685} \\
\cline{2-7}
\hline
\end{tabular}}
\label{tab:generalization_MLLM}
\end{table}

\begin{table}[t]
\caption{Computational efficiency comparison on VideoLLaMA2. Time is measured in seconds per sample. Best results are in \textbf{bold}, second-best are \underline{underlined}.}
\centering
\small
\setlength{\tabcolsep}{5pt}
\begin{tabular}{l|l|ccc}
\hline
Dataset & Method & Runtime$\downarrow$ & Acc2$\uparrow$ & MAE$\downarrow$ \\
\hline
\multirow{3}{*}{MOSI} 
& UBM & \textbf{15.35} & 86.1 & 0.716 \\
& VideoLLaMA2 \cite{damonlpsg2024videollama2} & 1046.23 & \underline{88.3} & \textbf{0.575} \\
& \textbf{U-ACS} & \underline{520.41} & \textbf{90.5} & \underline{0.625} \\
\hline
\multirow{3}{*}{SIMS} 
& UBM & \textbf{18.76} & 80.5 & 0.409 \\
& VideoLLaMA2 \cite{damonlpsg2024videollama2} & 1351.04 & \underline{81.0} & \underline{0.400} \\
& \textbf{U-ACS} & \underline{710.71} & \textbf{85.6} & \textbf{0.392} \\
\hline
\end{tabular}
\label{tab:videollama2_efficiency}
\end{table}

\subsection{Study on Hyperparameters}\label{subsec:Hyperparameters}
To analyze the impact of the threshold weighting hyperparameter $\lambda$ from Equation~\ref{eq:model_threshold}, we conducted a sensitivity analysis on MOSI and SIMS datasets. The hyperparameter $\lambda$ governs the uncertainty threshold, directly controlling how many samples are escalated to the MLLM. A lower $\lambda$ value results in a lower uncertainty threshold, causing more samples to be processed by the MLLM, while a higher $\lambda$ leads to a higher threshold, allowing the small model to handle more samples.

\begin{figure}[t]
    \subfigure[Acc2]{ 
        \begin{minipage}[t]{0.23\textwidth} 
        \centering
            \includegraphics[width=\linewidth]{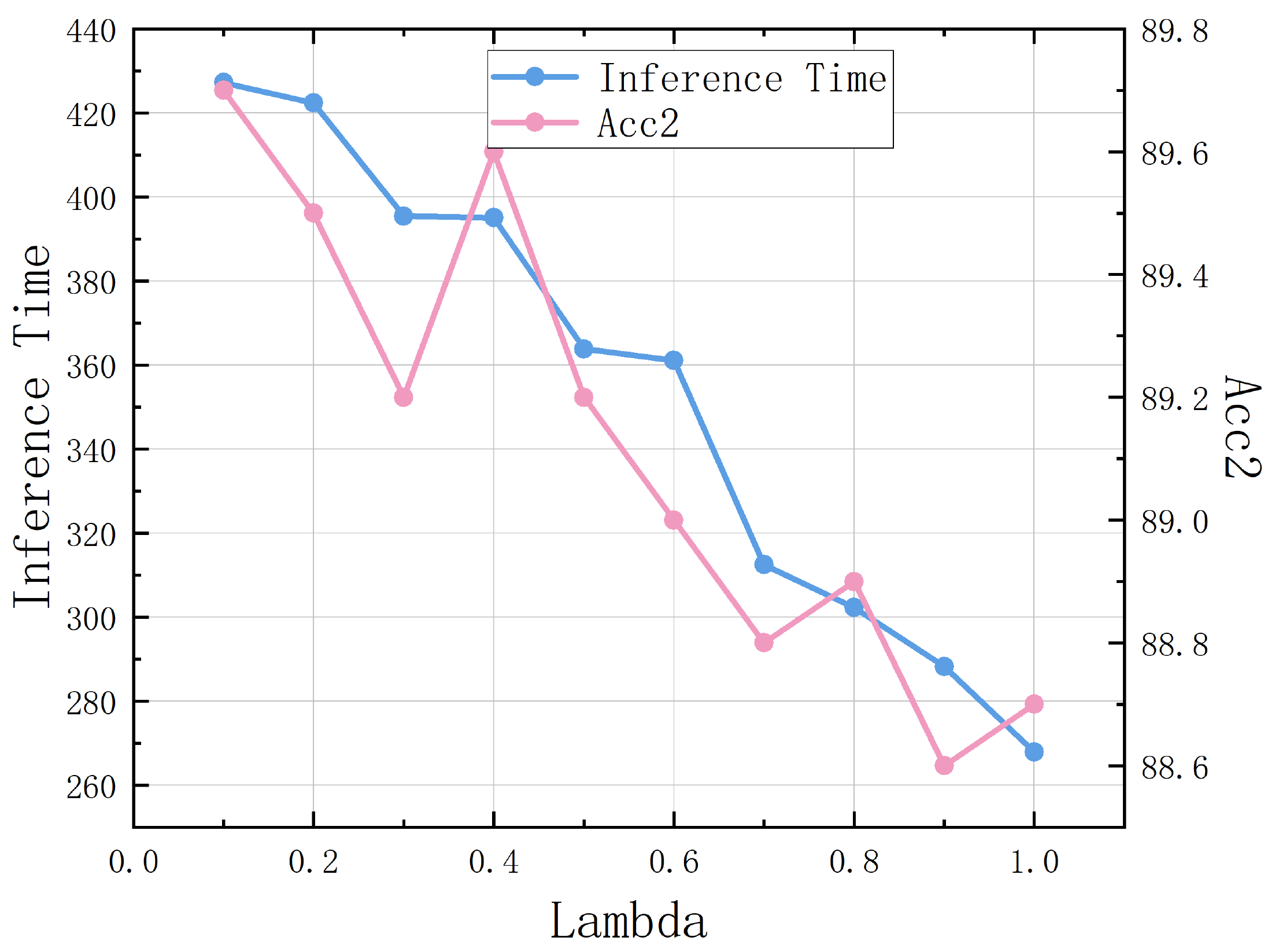}
            \label{fig:acc2:subfig:mosi}
        \centering
            \includegraphics[width=\linewidth]{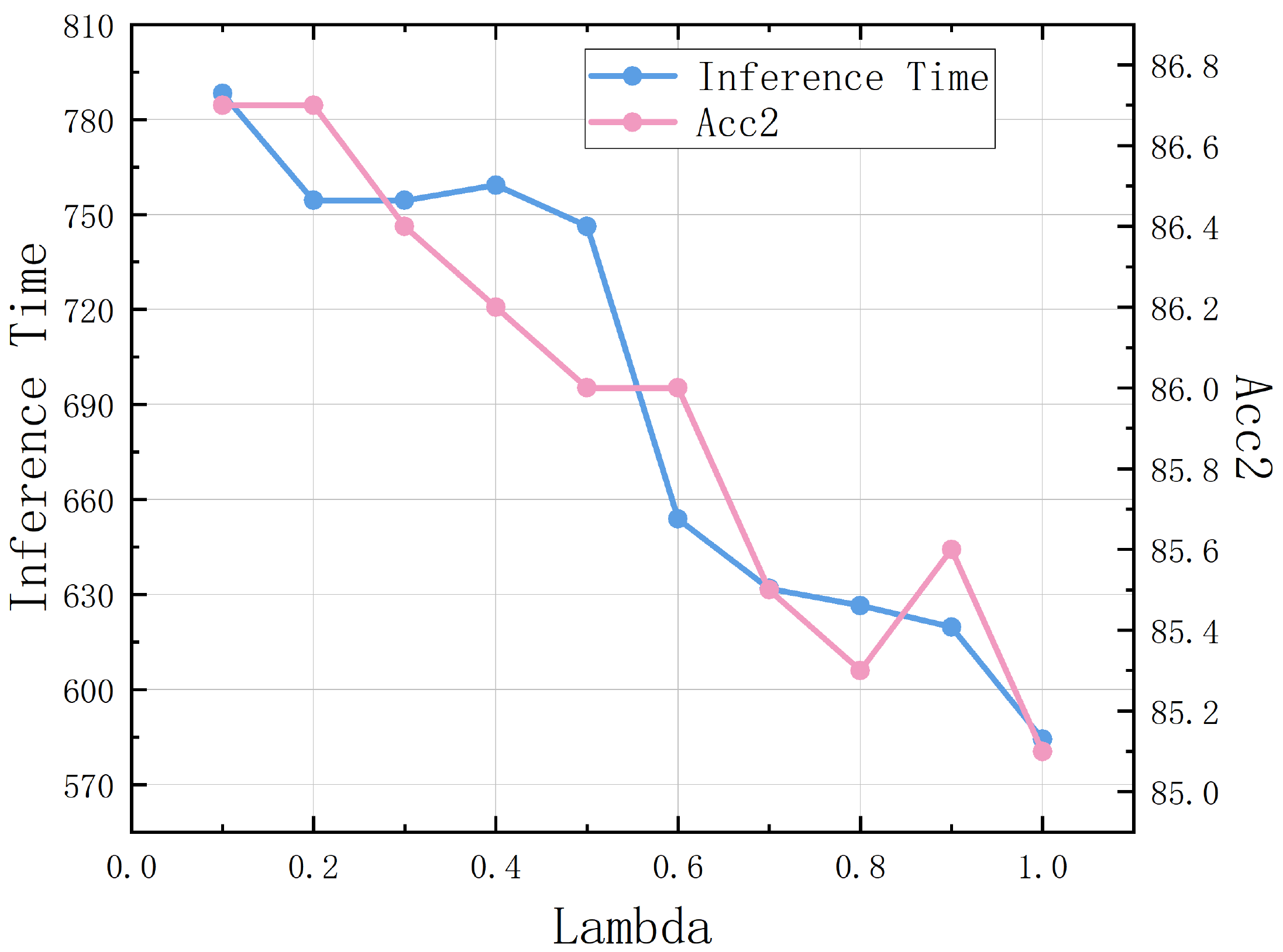}
            \label{fig:acc2:subfig:sims}
    \end{minipage}}%
    \subfigure[MAE]{ 
        \begin{minipage}[t]{0.23\textwidth}
        \centering
            \includegraphics[width=\linewidth]{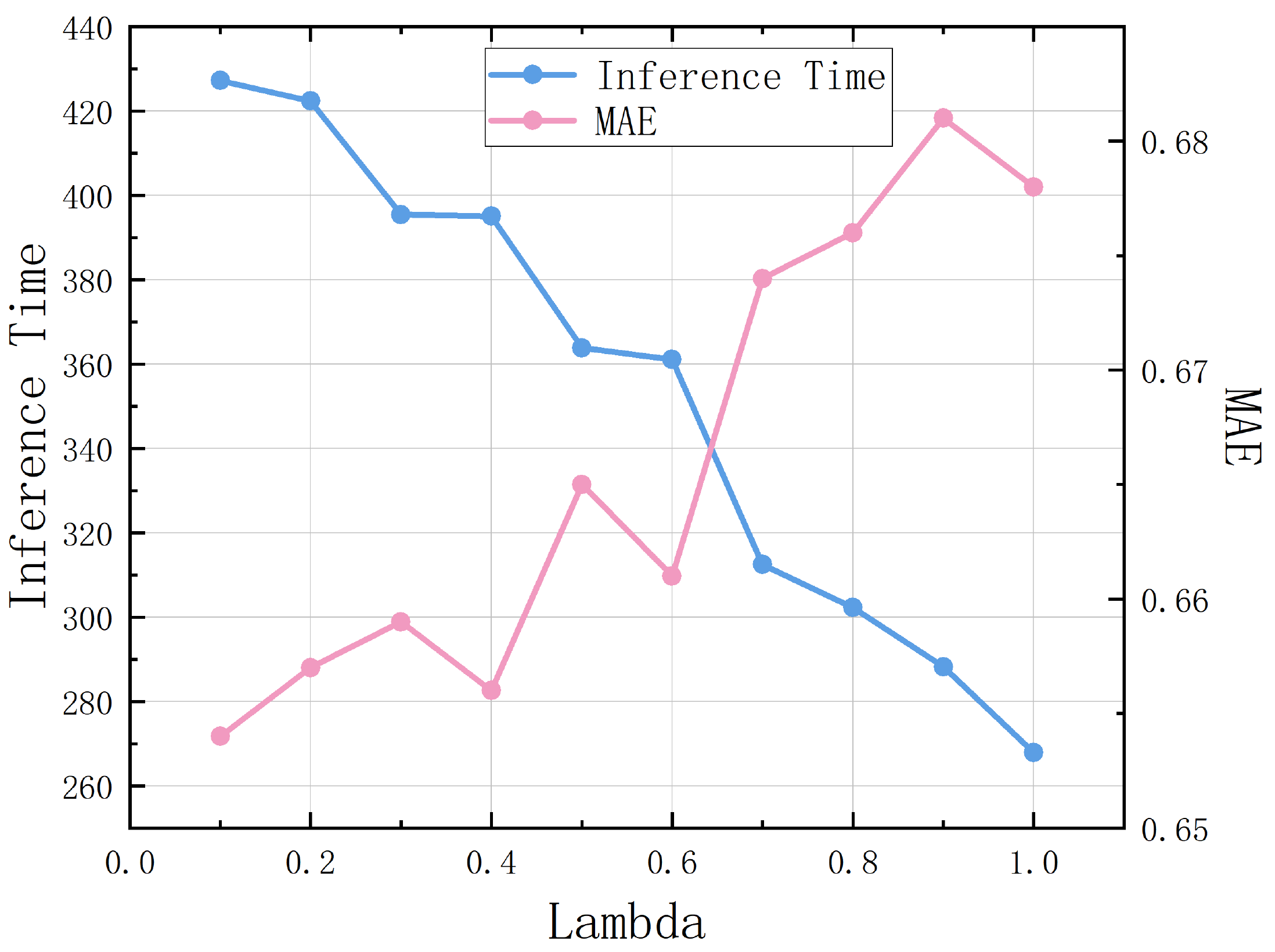} 
            \label{fig:mae:subfig:mosi}
        \centering
            \includegraphics[width=\linewidth]{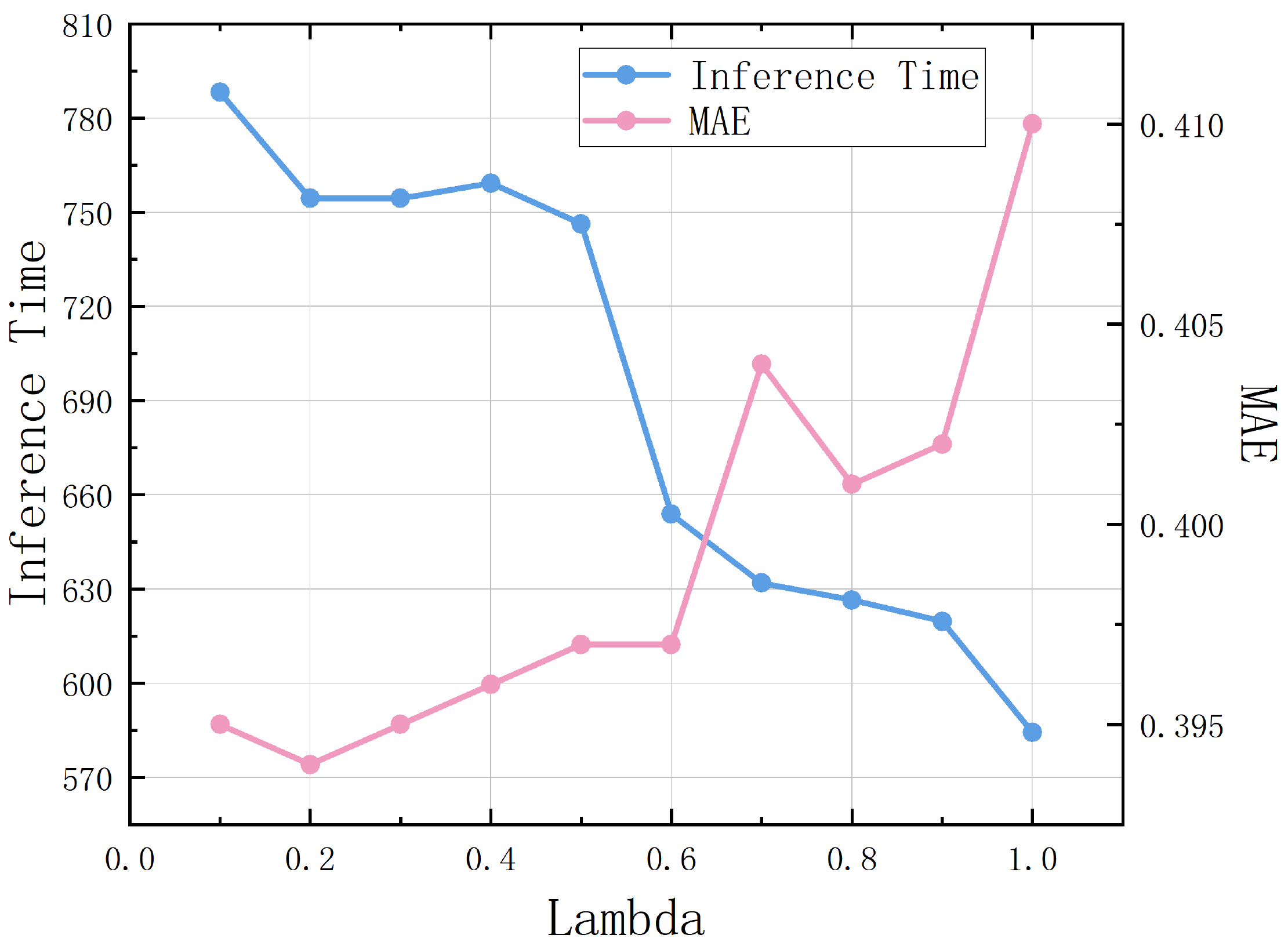} 
            \label{fig:mae:subfig:sims}
    \end{minipage}} 
    \caption{Impact of the threshold weighting parameter $\lambda$ on inference time and two metrics (Acc2 and MAE). Top row: MOSI. Bottom row: SIMS.}
\label{fig:lambda}
\vspace{-0.1cm}
\end{figure}

Figure \ref{fig:lambda} illustrates the trade-off between inference time and performance (Acc2 and MAE) as $\lambda$ varies from 0.1 to 1.0. The results reveal a clear trend:
At lower values of the hyperparameter $\lambda$, more samples are escalated to the MLLM, meaning the MLLM needs significantly more time to infer these samples. Yet the MLLM is endowed with more robust inference capabilities, which in turn enables the system to deliver better performance on the two metrics (Acc2 and MAE).
Thus, the primary factor affecting the system’s efficiency is the number of samples inferred by the MLLM. By adjusting the value of $\lambda$, we can find the optimal balance between inference time and model performance.

In Subsection \ref{sec:efficiency}, we elaborate on the efficiency of our system and demonstrate the trade-off between inference time and model performance. The hyperparameter $\lambda$ is therefore employed to identify an optimal value that maximizes the overall system performance. Consequently, when selecting hyperparameters, we need to ensure that the model achieves optimal performance across the target metrics. Therefore, for the main experiments in this study, we set the hyperparameter $\lambda$ to 0.1 for the MOSI dataset and to 0.2 for the SIMS dataset, respectively.

\subsection{Study on Uncertainty Distributions}\label{subsec:threshold}

\begin{figure}[t]
    \subfigure[UBM]{ 
        \label{fig:mosiexp:subfig:a}
        \begin{minipage}[b]{0.23\textwidth} 
        \centering
            \includegraphics[width=\linewidth]{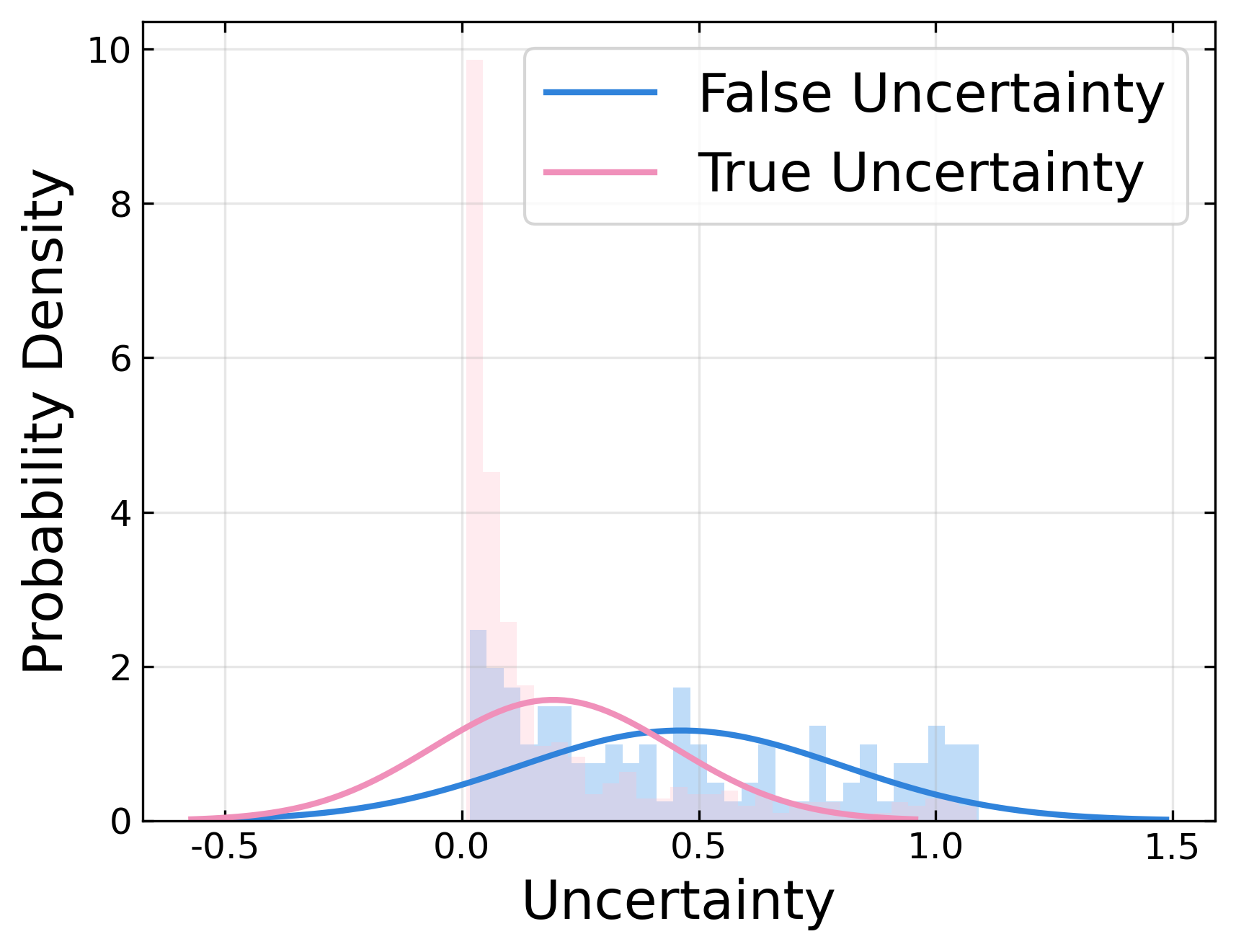} 
    \end{minipage}}%
    \subfigure[MLLM]{ 
        \label{fig:mosiexp:subfig:b}
        \begin{minipage}[b]{0.23\textwidth}
        \centering
            \includegraphics[width=\linewidth]{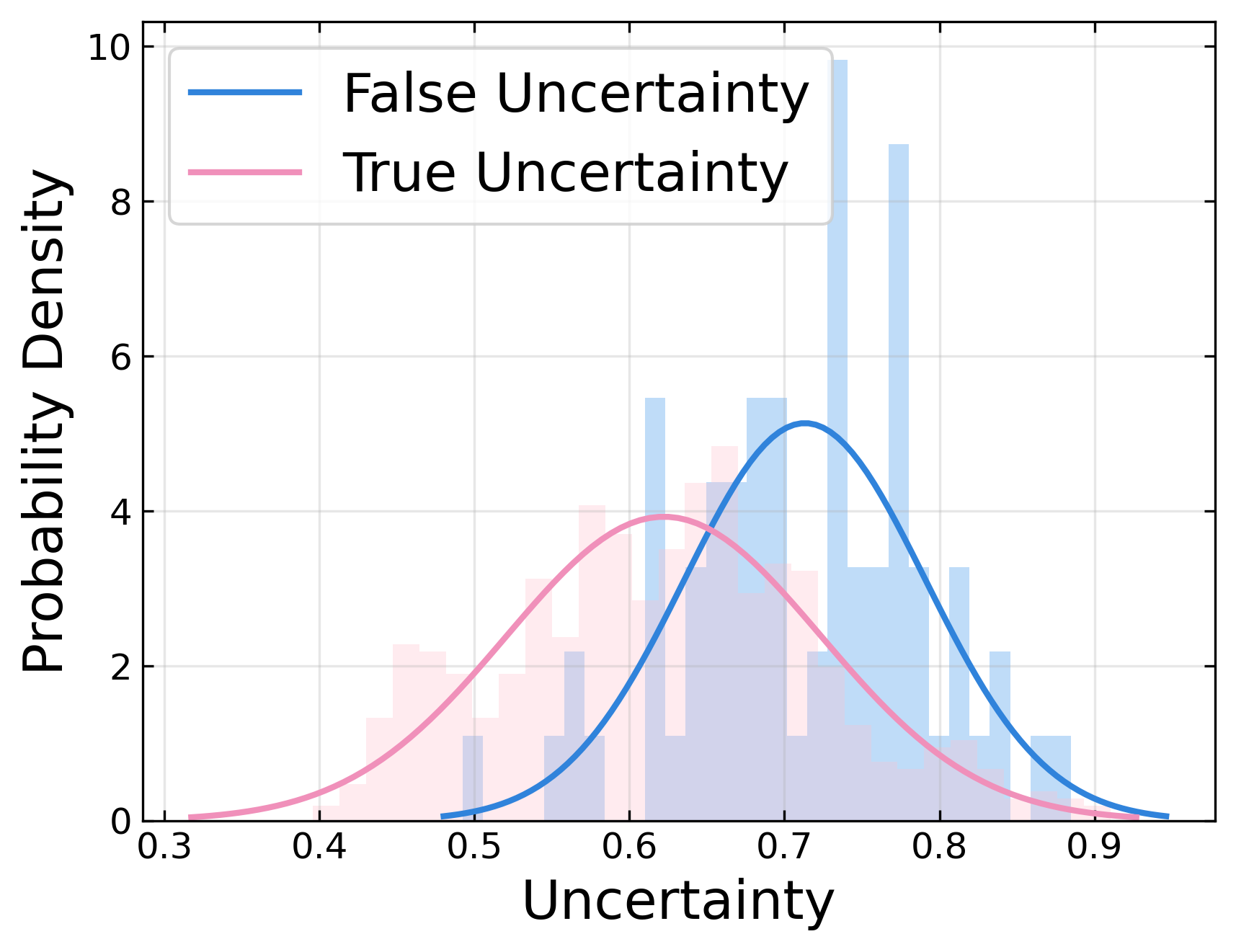} 
    \end{minipage}} 
    \vspace{-0.2cm}
    \caption{ The distribution of uncertainties on MOSI.}
\label{fig:mosiexp}
\vspace{-0.1cm}
\end{figure}

To determine optimal uncertainty thresholds $\tau_1$ and $\tau_2$, We fit the uncertainty of correctly and incorrectly predicted samples to a Gaussian distribution. 
Figure \ref{fig:mosiexp} shows the uncertainty distributions for correct and incorrect predictions on MOSI testing set.

Although there is an overlap in the uncertainties of some samples, the uncertainty distributions of correctly predicted samples and incorrectly predicted samples on the MOSI dataset still show significant differences. 
The visualization indicates the effectiveness of the proposed uncertainty estimation method (especially for UBM whose uncertainty is hard to estimate). 
The medians of the two Gaussian distributions exhibit a significant difference, indicating that predicted samples and incorrectly predicted samples have distinct separability in terms of numerical features. Thus, we can optimize the threshold to find an optimal dividing point, which can maximizes the separation between predicted samples and incorrectly predicted samples, thereby achieving effective classification of predicted samples and incorrectly predicted samples.

\begin{figure}[t]
\centering
\includegraphics[width=0.95\linewidth]{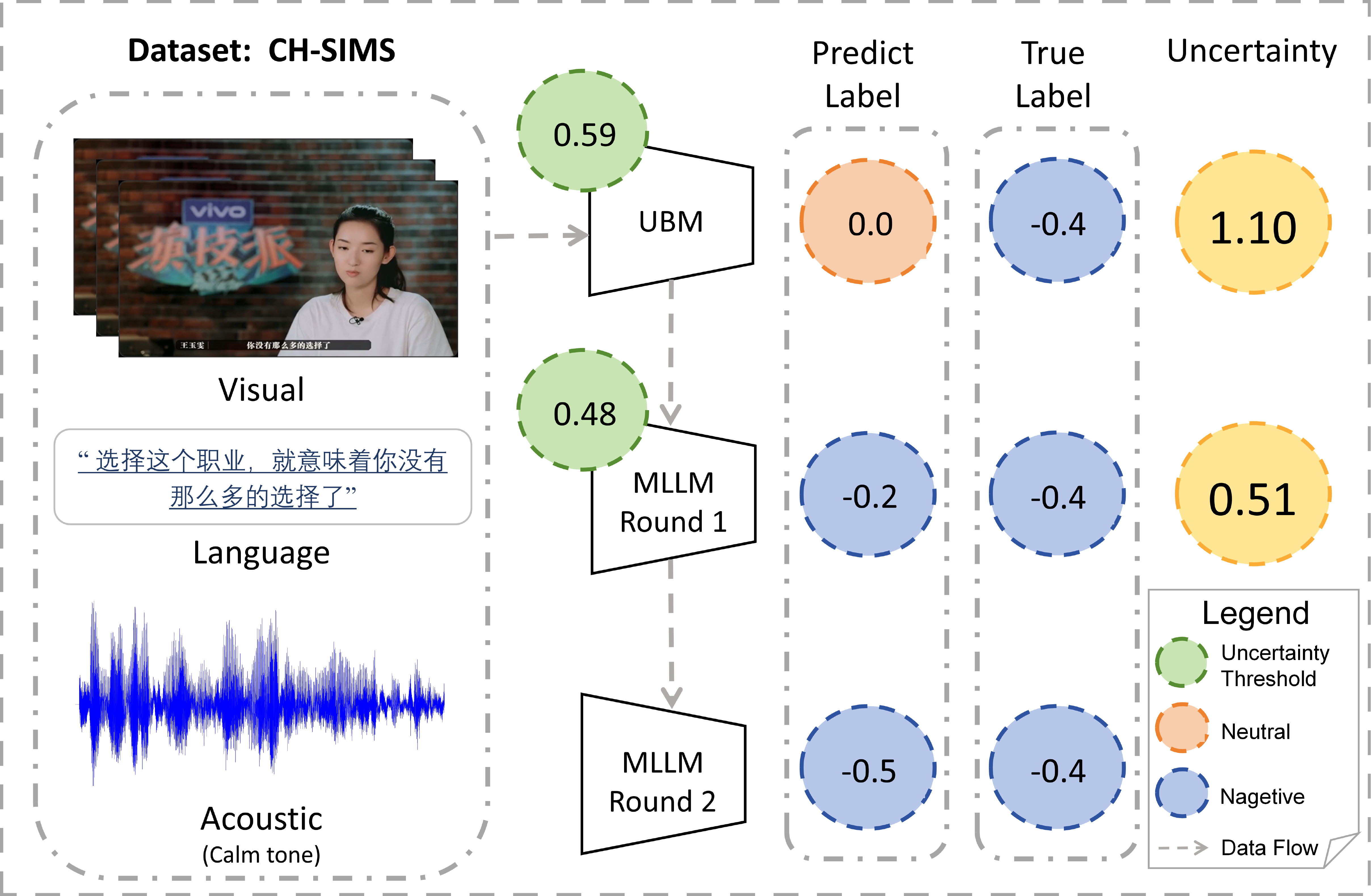}
\vspace{-0.1cm}
\caption{Case study on a challenging sample. UBM predicts neutral with high uncertainty triggering escalation. The MLLM first predicts with a uncertainty exceeding the threshold. A second MLLM run is activated for cross-verification, yielding a final and more precise prediction.}
\label{fig:case_study}
\vspace{-0.2cm}
\end{figure}

\subsection{Case Study}
In this subsection, to illustrate the effectiveness of our strategy, we present a case study on a challenging sample from the SIMS dataset. Figure \ref{fig:case_study} demonstrates how our framework processes a complex sentiment analysis case where individual models struggle but collaboration system succeeds.

The selected sample contains ambiguous multimodal cues: nuanced textual content discussing career limitations, subtle facial expressions, and calm acoustic tone. This complexity causes the UBM to predict neutral sentiment ($\hat{y}_s = 0.0$) with high uncertainty ($u_s = 1.10 > \tau_1 = 0.59$), correctly triggering escalation. The MLLM provides better polarity detection ($\hat{y}_l = -0.2$) but still exhibits uncertainty above the threshold ($u_l = 0.51 > \tau_2 = 0.48$), leading to collaborative processing. Since the predictions have conflicting polarities (neutral vs. negative), the cross-verification mechanism is activated, prompting a second MLLM inference that successfully achieves final prediction ($\hat{y}_f = -0.5$) with improved accuracy relative to the ground truth ($-0.4$).

This case demonstrates three key advantages. Firstly, uncertainty-guided routing prevents incorrect fast-path decisions when the baseline model faces ambiguous inputs. Secondly, the framework effectively handles polarity conflicts between models through cross-verification. Thirdly, the collaborative approach achieves superior accuracy on challenging samples .

\subsection{Ablation Studies}
\begin{table}[t]
\caption{Ablation experiments on MOSI and SIMS. Best results are in \textbf{bold}, second-best are \underline{underlined}.}
\centering
\small
\setlength{\tabcolsep}{4pt}
\begin{tabular}{l|l|ccc}
\hline
Dataset & Description & Acc7$\uparrow$  & Acc2$\uparrow$  & MAE$\downarrow$  \\
\hline
\multirow{5}{*}{MOSI} 
& W/o Cross-Verification & 50.4 & \underline{90.4} & 0.631 \\
& W/o Uncertainty 
& \textbf{51.7} & 89.6 & \underline{0.600} \\
& Prediction-Truth Difference & 46.7 & 87.0 & 0.649 \\
& Ensemble Variance & 46.9 & 88.7 & 0.632 \\
& \textbf{U-ACS} & \underline{51.2} & \textbf{91.8} & \textbf{0.585} \\
\hline
Dataset & Description & Acc3$\uparrow$  & Acc2$\uparrow$  & MAE$\downarrow$  \\
\hline
\multirow{5}{*}{SIMS} 
& W/o Cross-Verification & \underline{72.9} & \underline{85.1} & \underline{0.358} \\
& W/o Uncertainty 
& 70.7 & 83.6 & 0.357 \\
& Prediction-Truth Difference & 70.2 & 81.2 & 0.407 \\
& Ensemble Variance & 64.6 & 78.1 & 0.398 \\
& \textbf{U-ACS} & \textbf{73.1} & \textbf{85.8} & \textbf{0.354} \\
\hline
\end{tabular}
\label{tab:ablation_results}
\end{table}

We conduct comprehensive ablation studies to validate each component of our framework. Table \ref{tab:ablation_results} and Table \ref{tab:mosei_ablation} presents the results:



\textbf{(1) Uncertainty Routing Mechanism:} In `W/O Uncertainty', we randomly select identical samples for processing by the MLLM, rather than selecting samples based on uncertainty.  Removing the uncertainty routing mechanism significantly degrades performance (the Acc2 drops by 2.2 points and MAE also drops significantly on both datasets). The Acc5 also decreases by 3.1 points on SIMS, demonstrating the critical importance of sample routing mechanism. This is because uncertainty can reflect the confidence of model prediction, and samples of high uncertainty are often hard samples that should be processed more advanced models.

\textbf{(2) Cross-verification Strategies:} Although only a few samples will activate the cross-verification mechanism, it proves essential for optimal performance. Without it, Acc2 decreases by 1.4 points on MOSI and 0.7 points on SIMS, while MAE also deteriorates significantly, confirming the effectiveness of cross-verification. This is because cross-verification can effectively handle the hardest samples that cannot be well identified by individual models.

\textbf{(3) Uncertainty Quantification Methods:}
Our entropy-based approach substantially outperforms alternative uncertainty estimation methods. Compared to prediction-truth difference, we achieve 4.8-point improvements in Acc2 on MOSI and 4.6-point improvements on SIMS, and achieves 0.014 decrease in MAE and a 2.0-point improve in Acc7 on MOSEI.
Ensemble variance performs even worse, particularly on SIMS with a 7.7-point Acc2 gap and 9.2-point Acc5 gap. These results confirm that our single-pass entropy computation provides both efficiency and effectiveness for uncertainty estimation, which cleverly circumvents the issue of difficulty in computing uncertainty for regression task via transforming regression task into classification task.

\begin{table}[t]
\caption{Ablation experiments on MOSEI use HumanOmni as the MLLM. Best results are in \textbf{bold}, second-best are \underline{underlined}.}
\centering
\small
\begin{tabular}{l|l|ccc}
\hline
Dataset & Description & Acc7$\uparrow$ & Acc2$\uparrow$ & MAE$\downarrow$ \\
\hline
\multirow{3}{*}{MOSEI}
& PTD & 54.4 & \underline{84.7} & 0.520 \\
& EV & \underline{55.2} & \textbf{86.4} & \underline{0.516} \\
& \textbf{CBE} & \textbf{55.4} & \textbf{86.4} & \textbf{0.506} \\
\hline
\end{tabular}
\label{tab:mosei_ablation}
\end{table}



\subsection{Data Processing Volume in Different Components}
Tables~\ref{tab:humanomni_data} illustrate the actual volume of data processed across different components and datasets in our experiments. Owing to the varying levels of difficulty in sample processing across different datasets, the proportions of data routed to large and small models for processing also differ accordingly.

\begin{table}[t]
\caption{Data Processing Volume (DPV) across different components of U-ACS. The MLLM utilized in this table is HumanOmni.}
\centering
\small
\setlength{\tabcolsep}{5pt}
\begin{tabular}{c|c|l|c}
\hline
Dataset & Test set size & Method & DPV \\
\hline
\multirow{2}{*}{MOSI} 
& \multirow{2}{*}{686}
& UBM & 447  \\
& & HumanOmni & 239  \\
\hline
\multirow{2}{*}{MOSEI} 
& \multirow{2}{*}{4659}
& UBM & 1984  \\
& & HumanOmni & 2675  \\
\hline
\multirow{2}{*}{SIMS} 
& \multirow{2}{*}{457}
& UBM & 278  \\
& & HumanOmni & 179 \\
\hline
\end{tabular}
\label{tab:humanomni_data}
\end{table}


\section{Conclusion}\label{sec:Conclusion}

We propose a novel method named U-ACS to combine UBM with MLLMs to address the imbalance between performance and efficiency in MSA.
U-ACS explicitly computes the uncertainty of UBM and MLLM, designing an uncertainty-driven cascade mechanism to select samples yielding high predictive uncertainty for processing by the MLLM. We also propose advanced collaboration strategies to handle the hardest samples.
U-ACS significantly improves the efficiency of MLLM while ensuring model performance across multiple datasets. Moreover, we show that the proposed strategy can generalize to different MLLMs. 

\ifCLASSOPTIONcaptionsoff
  \newpage
\fi

\bibliographystyle{IEEEtran}


%

%

\bibliography{sample-base}

\end{document}